\documentclass{article}
\usepackage{float}
\usepackage{authblk}
\usepackage{algorithm}
\usepackage{algorithmicx}  
\usepackage{algpseudocode}  
\usepackage{setspace}
\usepackage{bm}
\usepackage[draft]{hyperref}
\usepackage{url}
\usepackage{amsmath}
\usepackage{stfloats}
\usepackage{mathrsfs}
\usepackage{dcolumn}
\usepackage{subfigure}  
\usepackage{multirow}
\usepackage{graphics}
\usepackage{graphicx}
\usepackage{geometry}
\usepackage{amsfonts} 
\usepackage{amssymb}
\usepackage{amsthm}
\usepackage{longtable}
\usepackage{booktabs}
\title{\textbf{Smooth Q-learning: Accelerate Convergence of Q-learning Using Similarity}}

\author[a,b]{Wei Liao}
\author[a,b]{Xiaohui Wei \thanks{Corresponding author: wei\_xiaohui@nuaa.edu.cn}}
\author[c]{Jizhou Lai}
\affil[a]{\footnotesize{ Key laboratory of Fundamental Science for National Defense-Advanced Design Technology of Flight Vehicle, 
Nanjing University of Aeronautics and Astronautics, Nanjing, Jiangsu, China }}
\affil[b]{\footnotesize{State Key Laboratory of Mechanics and Control of Mechanical Structures, 
Nanjing University of Aeronautics and Astronautics, Nanjing, Jiangsu, China}}
\affil[c]{\footnotesize{College of Automation Engineering, Nanjing University of Aeronautics and Astronautics, Nanjing, Jiangsu, China}}

\date{}

\begin{document}
\maketitle
\begin{abstract}
    An improvement of Q-learning is proposed in this paper. It is different from classic Q-learning in that the similarity between different states and 
    actions is considered in the proposed method. During the training, a new updating mechanism is used, 
    in which the Q value of the similar state-action pairs are updated synchronously. The proposed method
    can be used in combination with both tabular Q-learning function and deep Q-learning. And the results of numerical examples illustrate that 
    compared to the classic Q-learning, the proposed method has a significantly better performance.
\end{abstract}
\quad \\
\textbf{Keywards:} Reinforcement learning; Q-learning; Similarity

\section{Introduction}
Reinforcement learning aims to learn good policies for sequential decision problems \cite{art2} \cite{art2.1}. 
By maximizing the cumulative future reward, the performance of policies enhance as the training steps increase. 
Q-learning \cite{art2.2} is set of the most popular model-free reinforcement learning algorithms.
During the training, 
the state-action value function, known as Q value, is updated and the action with the biggest Q value is choosen at every time step.

\par A key point in the design of a Q-learning algorithms is the choice of a structure to store estimates of qualities \cite{art3}. 
In the original Q-learning method, a tabular-value function is used to represent the Q value \cite{art3}. 
In the tabular-value function, one element represents the Q value of oen state-action pair \cite{art1}. 
In paper \cite{art19}, a box-pushing problem is solved via the Q-learning method with tabular-value function. 
And in some examples of book \cite{art21}, tabular-value function is used and excellent results were achieved.
However, the size of the table may be considerable 
because of the excessive amount of memory needed to store the table \cite{art20}. In order to deal with the continuous state space, some 
approximation methods are taken. The linear combination of some linear function of state is used to express the Q value in \cite{art4}.
With the development of deep learning, neural network (NN) with several layers of
nodes, called deep Q-network (DQN), is used to build up progressively more abstract representations of the Q value \cite{art5} \cite{art2}. 
For the past few years, with the going deep of the research work, some improved form of DNQ were proposed. 
In order to overcome the overestimation of action values caused by insufficiently flexible function approximation %7.Thrun and Schwartz, 1993
a new new mechanism of evaluating the action values was proposed in \cite{art7}. 
A dueling architecture was proposed in \cite{art8}, where the Q value estimates is represented as the combination of the state-value and the advantages for
each action. In order to make the most effective use of the replay memory for learning, instead of random sampling from the replay memory, paper \cite{art9}
proposed a new sampling method based on priority. Recently, the generative antagonistic
network (GAN) is also used in Q-learning, paper \cite{art18} proposed GAN Q-learning and analyzed its performance
in simple tabular environments.

\par Q-learning has been applied successfully in many engineering fields \cite{art10} \cite{art11} \cite{art22}. However, there are still some shortcomings. 
Tabular Q-learning can only solve the problems with limited and discrete states and actions, too large amounts of states or actions always
leads to an extended period training time. 
Although DQN can deal with the problems with continuous state space, it can 
only handle discrete action spaces. An obvious approach to adapting DQN to continuous
domains is to simply discretize the action space \cite{art12}. This approach has a limitation: 
The more parts the action space is discretized into, the more time is spent in training.
A truth is that, intuitively, taking similar actions at similar states always obtain roughly equal cumulative future reward. 
Based on this truth, an improvement of Q-learning is proposed in this paper: When a Q value of a state-action pair is updated, 
the Q value of the similar state-action pairs are updated synchronously.
It is worth pointing out that the proposed method can be used in combination with both tabular-value function and DQN.

\par The rest of the paper is organized as follows. Background
and some preliminaries are introduced in Section 2. 
Section 3 presents the details of the proposed method. 
Some numerical examples are taken to demonstrate the effectiveness of our method in Section 4.
The results are summarized in Section 5.

\section{Background}
\par Consider a standard reinforcement learning problem consisting of an agent interacting with an environment
$E$ in discrete timesteps. At each timestep $t$ the agent receives an observation $s_t \in \mathcal{S}$, takes
an action $a_t \in \mathcal{A}$ and receives a scalar reward $r_t$. 
\par An agent's behavior is defined by a policy, $\pi$, which maps from state space $\mathcal{S}$ to action space $\mathcal{A}$, 
$\pi: \mathcal{S}\to \mathcal{A}$. The environment $E$ can be modeled as a Markov
decision process (MDP) with a state space $\mathcal{S}$, an action space $\mathcal{A}$, a transition dynamics $s_{t+1} = f(s_t,a_t)$ 
and a reward function $r_t=r(s_t,a_t)$. The Q value function is used in many reinforcement learning algorithms. It describes the expected
return after taking an action $a$ in state $s$ and thereafter following policy $\pi$:
\begin{align}
  Q^{\pi}(s,a)=\sum_{t=0}^{\infty}\left[\gamma^t r_t | s_0=s,a_0=a,\pi   \right] 
\end{align}
where $\gamma\in [0,1]$ is a discount factor that trades off the importance
of immediate and later rewards. Q-learning aims to accurately estimate and maximize the Q value of each state-action pair and then 
an optimal policy is easily derived from the optimal values by selecting the highest valued
action in each state. 

\section{Method Details}
\subsection{Tabular-value Function}
\par For the reinforcement learning problems with limited states and actions, the Q value of each state-action pair can be stored
in a tabular-value function. In the classic Q learning method, after taking action $a_t$ in state $s_t$ and getting the immediate reward $r_t$
and resulting state $s_{t+1}$, the tabular-value function is updated as follows:
\begin{align}
  Q(s_t,a_t) 
  =Q(s_t,a_t)+\alpha \left[r_t+ \gamma \max_a Q(s_{t+1},a)-Q(s_t,a_t) \right]
\end{align}
where $\alpha$ is a scalar step size. 
\par The similarity between different states and actions can be measured by norm. 
In our method, the Q value of a state-action pair $(s,a)$ is updated synchronously if it satisfy
\begin{align}
  \left\| s-s_t \right\| \leq \delta_s 
\end{align}
or 
\begin{align}
  \left\| a-a_t \right\| \leq \delta_a 
\end{align}
where $ \delta_s>0 $ and $\delta_a>0$ are two threshold values. 
The details are shown in 
Algorithm 1 and Algorithm 2, where the former takes the similarity of different states into consideration during the training and 
the similarity of different actions is taken into consideration in the latter, $\beta_s>0$ and $\beta_a>0$ are parameters of the algorithms named smooth rate,
which determine the updating step size of the Q value of similar state-action pairs.

\begin{algorithm}  
  \begin{algorithmic}
    \caption{State Space Smooth Tabular Q-learning}
    \State Initialize Q value function as $0$ for all state-action pairs;
    \State $s=s_0$;
    \For{$t=1,2,...$}
      \State With probability $\epsilon$ select a random action $a$, 
      \State otherwise $a=\arg \max_{a'} Q(s,a';\bm{\theta})$;
      \State Execute action $a$ and observe reward $r$ and observe new state $s_{next}$;
      \State $q=r+\gamma \max_{a'} Q(s_{next},a')$;
      \State $Q(s,a)=Q(s,a)+\alpha [q-Q(s,a)]$;
      \For{$s' \in \mathcal{S}$}
          \If{$\left\| s'-s \right\|\leq \delta_s$}
            \State $q'=(1-\beta_s)Q(s',a)+\beta_s q$;
            \State $Q(s',a)=Q(s',a)+\alpha \left[q'-Q(s',a)  \right] $;
          \EndIf
      \EndFor
      \State $s=s_{next}$;
    \EndFor
  \end{algorithmic}
\end{algorithm}

\begin{algorithm}  
  \begin{algorithmic}
    \caption{Action Space Smooth Tabular Q-learning}
    \State Initialize Q value function as $0$ for all state-action pairs;
    \State $s=s_0$;
    \For{$t=1,2,...$}
      \State Generate an equally distributed random real number $d$ uniformly distributed on the interval from 0 to 1;
      \State With probability $\epsilon$ select a random action $a$, 
      \State otherwise $a=\arg \max_{a'} Q(s,a';\bm{\theta})$;
      \State Execute action $a$ and observe reward $r$ and observe new state $s_{next}$;
      \State $q=r+\gamma \max_{a'} Q(s_{next},a')$;
      \State $Q(s,a)=Q(s,a)+\alpha [q-Q(s,a)]$;
      \For{$a' \in \mathcal{A}$}
          \If{$\left\| a'-a \right\|\leq \delta_a$}
            \State $q'=(1-\beta_a)Q(s,a')+\beta_a q$;
            \State $Q(s,a')=Q(s,a')+\alpha \left[q'-Q(s,a')  \right] $;
          \EndIf
      \EndFor
      \State $s=s_{next}$;
    \EndFor
  \end{algorithmic}
\end{algorithm}

\subsection{ Deep Q Network }
\par Due to the limitation of tabular-value function, DQN is widely used in engineering practice. 
For each time step, the DQN with parameters $\bm{\theta}$ of standard Q-learning method, say $Q(s,a;\bm{\theta})$, is updated as follows
\begin{align}
  \bm{\theta}=\bm{\theta}+\alpha \left[ Y-Q(s_t,a_t;\bm{\theta})  \right] \nabla_{\bm{\theta}}Q(s_t,a_t;\bm{\theta})
\end{align}
where
\begin{align}
  Y=r_{t+1} + \gamma \max_a Q(s_{t+1},a;\bm{\theta})
\end{align}

Neural networks can fit the continuous functions, therefore, the similarity of different states is naturally taken into 
consideration when the DQN is used to represent the Q value. However, the action space is still considered as discrete space 
in deep Q-learning and its improved versions. Based on the deep Q-learning proposed in [nature DQN], know as Nature DQN, 
and considered similarity between different actions, a new updating mechanism of DQN is introduced in this subsection, see Algorithm 3.

\begin{algorithm}  
  \begin{algorithmic}
    \caption{Smooth Deep Q-learning}
    \State Initialize replay memory $D$ to capacity $N$;
    \State Initialize state-action value function $Q$ with random parameters $\bm{\theta}$;
    \State Initialize target state-action value function $\hat{Q}$ with random parameters $\bm{\theta}^*=\bm{\theta}$;
    \State $s=s_0$;
    \For{$t=1,2,...$}
      \State With probability $\epsilon$ select a random action $a$, 
      \State otherwise $a=\arg \max_{a'} Q(s,a';\bm{\theta})$;
      \State Execute action $a$ and observe reward $r$ and new state $s_{next}$;
      \State Store transition $(s,a,r,s_{next})$ in $D$;
      \State Sample random minibatch of transitions $\left\{  (s_j,a_j,r_j,s_{next,j}) \right\}$ from $D$;
      \State $q_j=r_j+\gamma \max_{a'} \hat{Q}(s_{next,j},a';\bm{\theta^*})$;
      \State $l_0=\sum_j \left[ q_j-Q(s_j,a_j;\bm{\theta})  \right]^2$;
      \State $l_1=\beta \sum_j \left\{  \sum_{ \left\| a'-a_j \right\|\leq \delta_a,a'\neq a_j  } \left[ q_j-Q(s_j,a';\bm{\theta})  \right]^2    \right\}$
      \State Perform a gradient descent step on $l_0+l_1$ with respect to $\bm{\theta}$;
      \State $q=r+\gamma \max_{a'} Q(s_{next},a')$;
      \State $Q(s,a)=Q(s,a)+\alpha [q-Q(s,a)]$;
      \State Every $C$ steps reset $\bm{\theta}^*=\bm{\theta}$;
      \State $s=s_{next}$;
    \EndFor
  \end{algorithmic}
\end{algorithm}

\section{Numerical Examples}
\subsection{Smooth Tabular Q-learning}
\par Consider a MDP involved a 64x64 gridworld, the state of the agent is expressed as a position $(x,y)$ in the gridworld.  
For every time step, the action $(a_x,a_y)$ can be choosen from the set $\left\{ (-1,0),(0,1),(1,0),(0,-1)   \right\}$. 
The transition dynamics is 
\begin{align}
  \begin{cases}
    x_{t+1}=\min[ \max (x_t+a_x,0),63]\\
    y_{t+1}=\min[ \max (y_t+a_y,0),63]
  \end{cases}
\end{align}
The reward function is 
\begin{align}
  r_t=
  \begin{cases}
    100,x_t=63 \ \mathrm{and} \ y_t=63 \\
    -1, \mathrm{otherwise}
  \end{cases}
\end{align}

At the beginning of each training episode, the state of agent is set as $(0,0)$, and the training episode is end when the 
agent get the state $(63,63)$. The simulation results of Algorithm 1 with different $\beta_s$ and Algorithm 2 with different $\beta_a$ are shown in Figure 1 and Figure 2 respectively. 
During the simulations, the other parameters are set as follows
\begin{align*}
  &\epsilon=0.1\\
  &\alpha=0.1\\
  &\gamma=0.9\\
  &\delta_s=1
\end{align*}

\begin{figure}\centering
  \includegraphics[width=7cm]{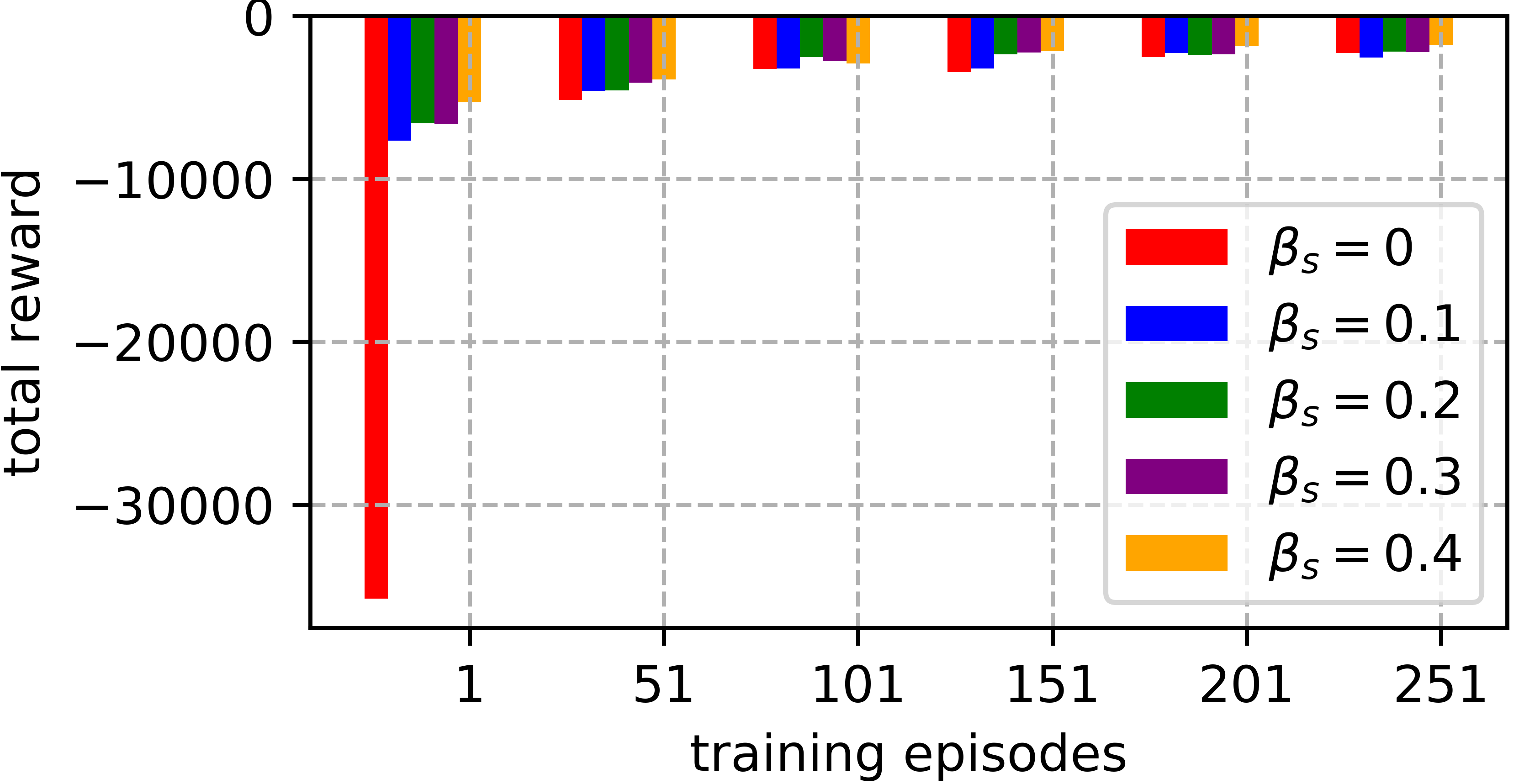}
  \caption{Simulation results of Algorithm 1.}
\end{figure}

\begin{figure}\centering
  \includegraphics[width=7cm]{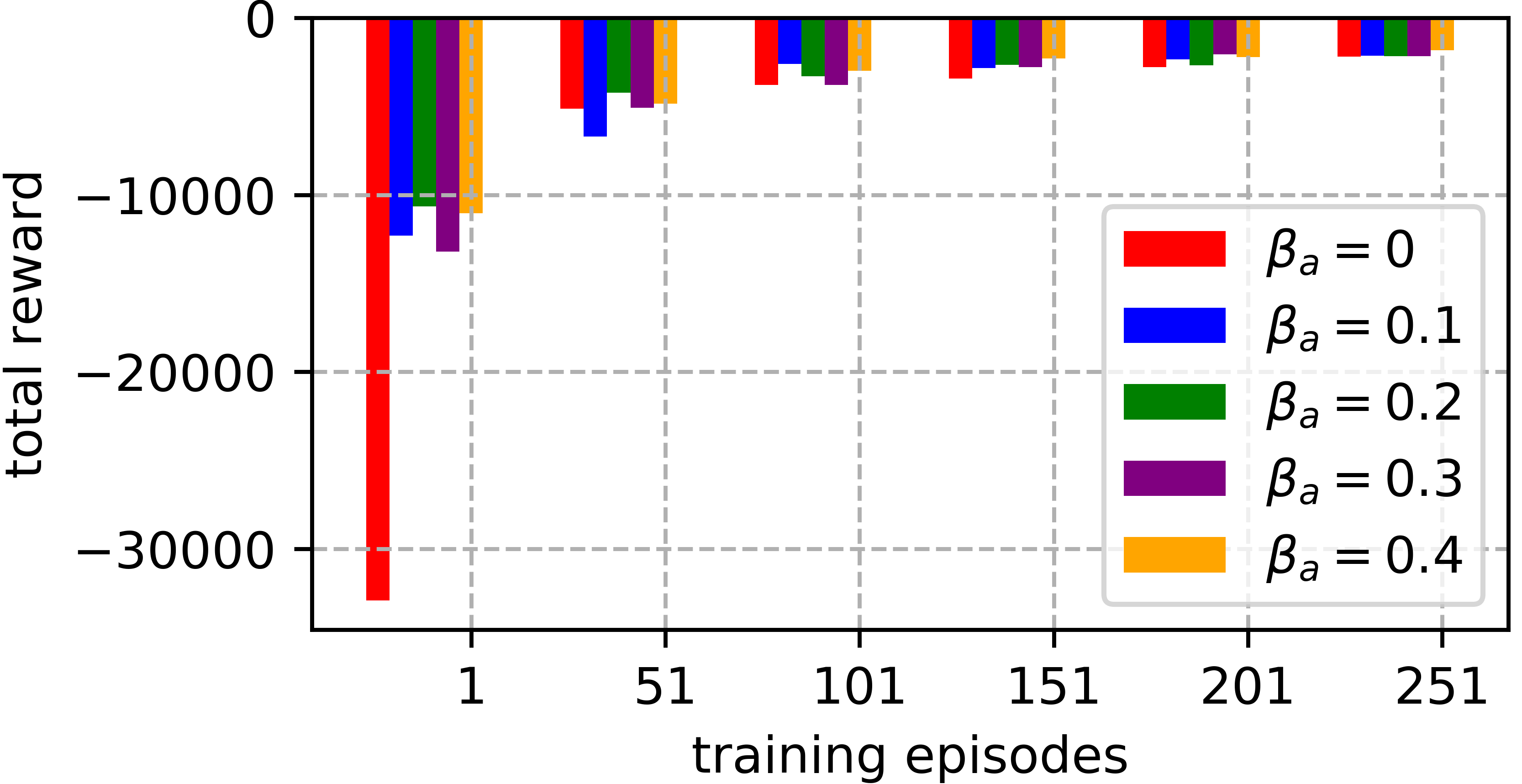}
  \caption{Simulation results of Algorithm 2.}
\end{figure}

It can be seen from Figure 1 and Figure 2 that with the consideration of the similarity between different 
states and actions, the Q value function has a significantly faster convergence speed.

\subsection{Smooth Deep Q-learning}
\par In this subsection, some classical RL examples with continuous action space are taken. The first is 
the Pendulum problem [gym website]. The transition dynamics is 
\begin{align}
  \begin{cases}
    &\omega_{t+1}=\min\{ \max\left[ \left( \omega_t-\frac{3g}{2l}\sin \theta_t+\frac{3u}{ml^2}  \right)0.05,-8  \right],8 \}\\
    &\theta_{t+1}=(\theta_t+0.05\omega_{t+1}+\pi)\%(2\pi)-\pi
  \end{cases}
\end{align}
\par where $\theta$ and $\omega$ are angle and angle velocity respectively. $g=10$, $m=1$ and $l=1$ are 
acceleration of gravity, mass and length respectively. The operator "\%" is modulo operator. 
$u \in [-2,2]$ is action. In order to adapt DQN to this problem, the action space is discretized into several parts, that is 
\begin{align}
  u \in \left\{ -2+\frac{4}{63}i| i=0,1,...,63  \right\}
\end{align}
And the reward function is 
\begin{align}
  r(\theta_t,\omega_t,u_t)=-\theta_t^2-\omega_t^2-0.001u_t^2
\end{align}

\par The second example is MountainCar. The transition dynamics is
\begin{align}
  \begin{cases}
    v_{t+1}=\min\left[  \max \left( v_t+0.0015u-\frac{\cos 3p_t}{400},-0.07  \right),0.07   \right]\\
    p_{t+1}=\min\left[  \max(p_t+v_{t+1},-1.2),0.6   \right]
  \end{cases}
\end{align}
Where $p$ and $v$ are position and velocity respectively, and $u\in [-1,1]$ is the action. Similar to the previous example, 
the action space is discretized into several parts
\begin{align}
  u \in \left\{ -1+\frac{2}{63}i| i=0,1,...,63  \right\}
\end{align}
And the reward function is 
\begin{align}
  r(p_t,v_t,u_t)=
  \begin{cases}
    &20p_{t+1}-0.1u^2, p_{t+1}<0.45\\
    &100-0.1u^2, p_{t+1}\geq 0.45
  \end{cases}
\end{align}

The simulation results of the two examples are shown in Figure 3 and Figure 4. It is clear that the method with a smooth rate 
greater than 0 (smooth Q-learning) does substantially better than the classic Q-learning.
\begin{figure}\centering
  \includegraphics[width=7cm]{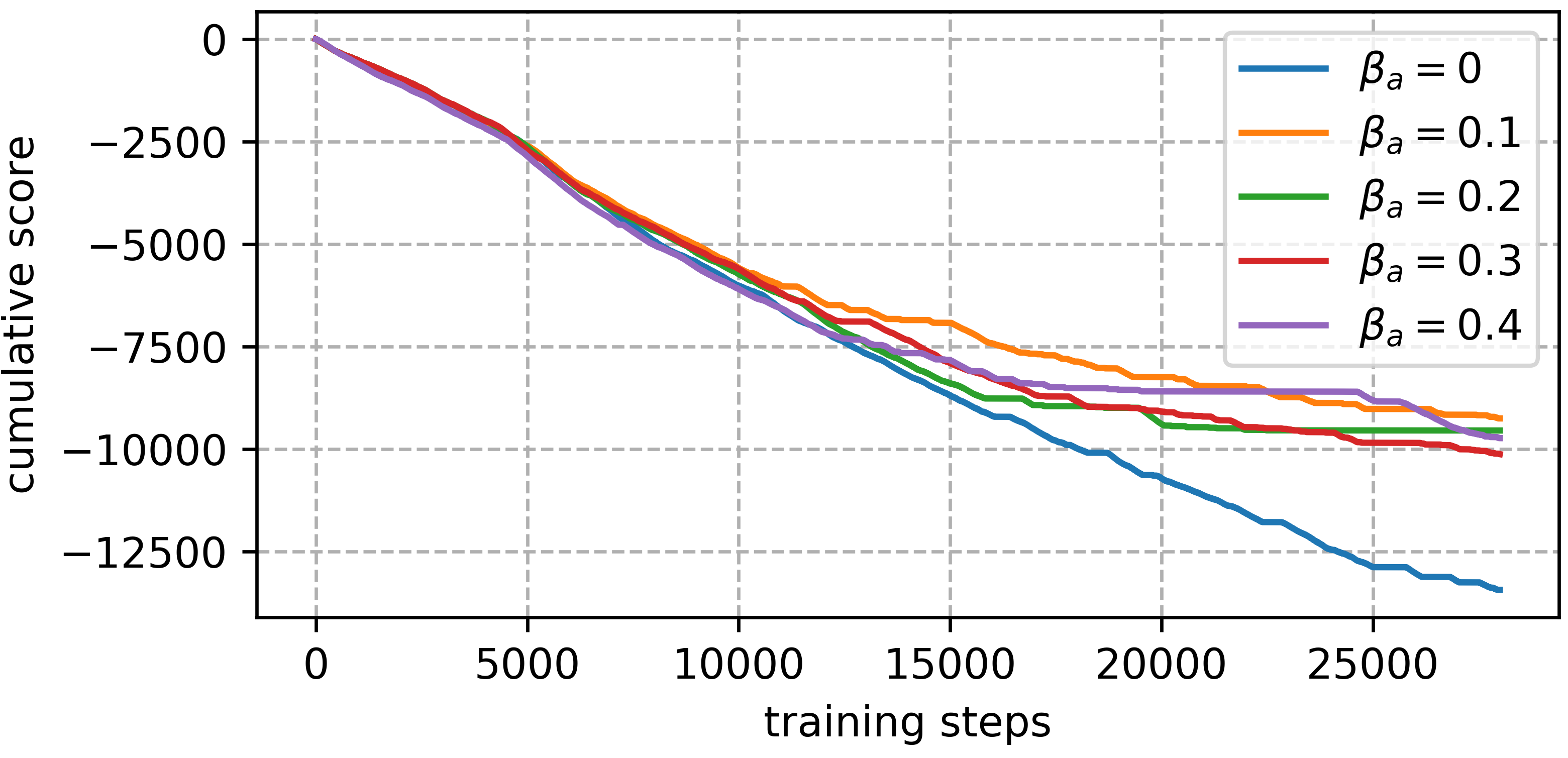}
  \caption{Simulation results of Pendulum problem.}
\end{figure}

\begin{figure}\centering
  \includegraphics[width=7cm]{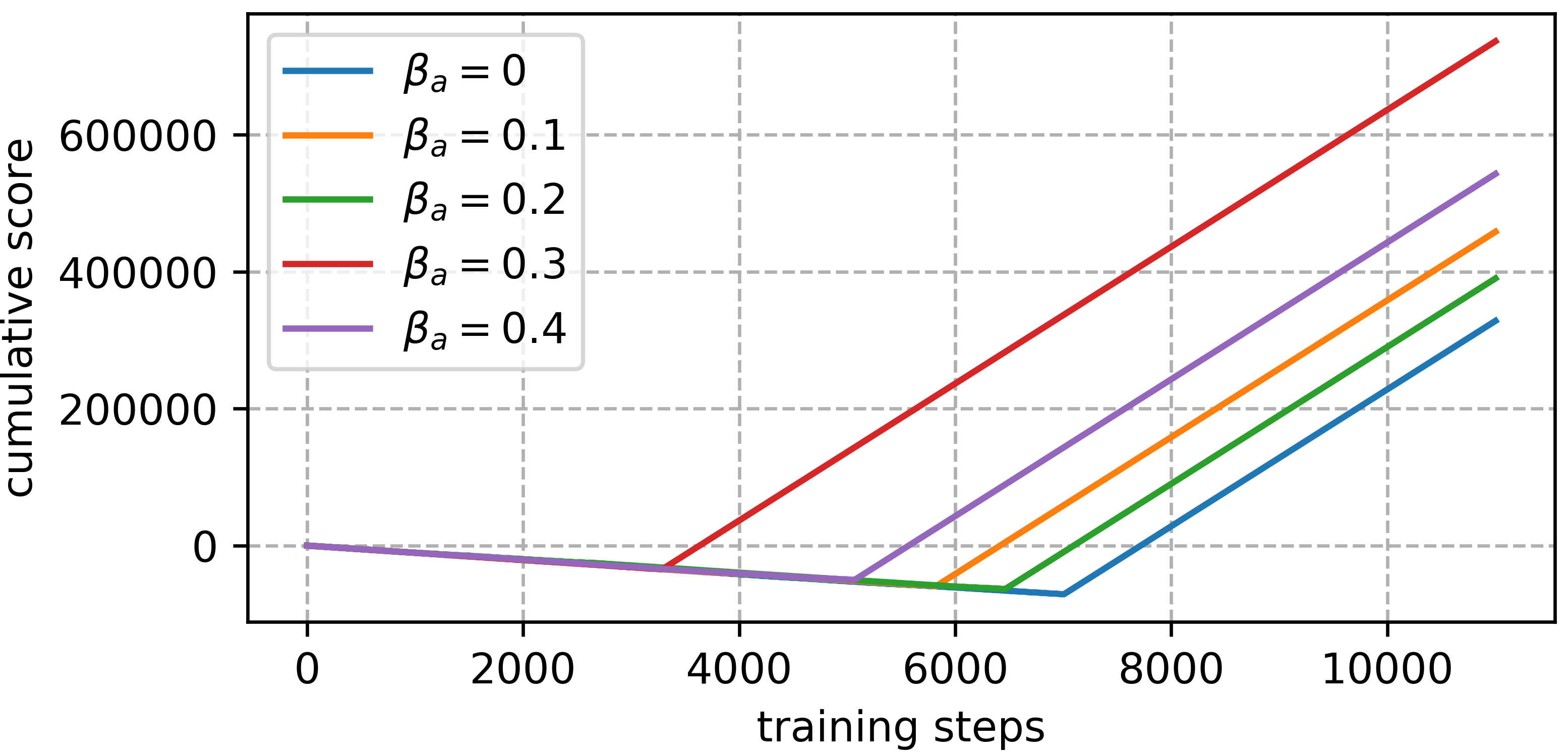}
  \caption{Simulation results of MountainCar problem.}
\end{figure}

\section{Conclusions}
\par This paper presents an improvement of Q-learning method, in which the similarity of different states or/and actions
is considered. Through a new updating mechanism, the convergence is obviously accelerated. Besides the chosen states and actions, 
the Q value of the similar state-action pairs are updated synchronously in every training step. 
The simulation results illustrate that, compared to the existing Q-learning method, the performance of the proposed method has remarkable advantage.
However, over the course of the study, some problems were found. For example, the convergence speed is not monotonely increasing along the smooth rate,
to determine the optimal smooth rate is necessary. The proposed method has a poor performance when it is adapted in the problems with high-dimensional continuous 
action space. These problems will be considered in our future
work.

\section*{Acknowledgements}
\par The authors gratefully acknowledge support from
National Defense Outstanding Youth Science Foundation
(Grant No. 2018-JCJQ-ZQ-053), and Research
Fund of State Key Laboratory of Mechanics and Control
of Mechanical Structures (Nanjing University of
Aeronautics and astronautics) (Grant No. MCMS-0217G01). Also, the authors would like to thank the
anonymous reviewers, associate editor, and editor for
their valuable and constructive comments and suggestions.

\bibliographystyle{unsrt}

% Loading bibliography database
\bibliography{mybib}

% \section*{Acknowledgements}
% \par The authors gratefully acknowledge support from National Defense Outstanding Youth Science Foundation (Grant No. 2018-JCJQ-ZQ-053), 
% and Central University Basic Scientific Research Operating Expenses Special Fund Project Support (Grant No. NF2018001).

\end{document}